\documentclass[letterpaper]{article} 
\usepackage{aaai24}  
\usepackage{times}  
\usepackage{helvet}  
\usepackage{courier}  
\usepackage[hyphens]{url}  
\usepackage{graphicx} 
\usepackage{natbib}  
\usepackage{caption} 
\usepackage{algorithm}
\usepackage{algorithmic}
\usepackage{footnote}
\usepackage{amssymb}
\usepackage{mathtools}
\usepackage{multirow}
\usepackage{amsmath}
\usepackage{amsthm}
\usepackage{graphicx}
\usepackage{subfigure}
\usepackage{caption}
\usepackage{subcaption}
\usepackage{enumitem}
\setlist{nosep}
\usepackage{booktabs}    

\usepackage[T1]{fontenc}
\theoremstyle{definition} 
\newtheorem{definition}{Definition}
\newcommand{\retro}{retrosynthesis }

\usepackage{newfloat}
\usepackage{listings}
\DeclareCaptionStyle{ruled}{labelfont=normalfont,labelsep=colon,strut=off} 
\floatstyle{ruled}
\newfloat{listing}{tb}{lst}{}
\floatname{listing}{Listing}
\title{RetroOOD: Understanding Out-of-Distribution \\ Generalization in Retrosynthesis Prediction}
\author{
    Yemin Yu\textsuperscript{\rm 1,5}\equalcontrib,
    Luotian Yuan\textsuperscript{\rm 2}\equalcontrib,
    Ying Wei\textsuperscript{\rm 3}\thanks{Corresponding author.},
    Hanyu Gao \textsuperscript{\rm 4},
    Xinhai Ye \textsuperscript{\rm 5},
    Zhihua Wang \textsuperscript{\rm 5},
    Fei Wu \textsuperscript{\rm {2,5}}
}
\affiliations{
    \textsuperscript{\rm 1}City University of Hong Kong,
    \textsuperscript{\rm 2}Zhejiang University,
    \textsuperscript{\rm 3}Nanyang Technological University \\
    \textsuperscript{\rm 4}Hong Kong University of Science and Technology, 
    \textsuperscript{\rm 5}Shanghai Institute for Advanced Study of Zhejiang University \\
    yeminyu2-c@my.cityu.edu.hk, 12221240@zju.edu.cn, ying.wei@ntu.edu.sg, hanyugao@ust.hk \\
    yexinhai@zafu.edu.cn, \{zhihua.wang, feiwu\}@zju.edu.cn
}

\begin{document}

\maketitle

\begin{abstract}
Machine learning-assisted retrosynthesis prediction models have been gaining widespread adoption, though their performances oftentimes degrade significantly when deployed in real-world applications embracing out-of-distribution (OOD) molecules or reactions. 
Despite steady progress on standard benchmarks, our understanding of existing~retrosynthesis prediction models under the premise of~distribution shifts remains stagnant.
To this end, we first formally sort out two types of distribution shifts in retrosynthesis prediction and construct two groups of benchmark datasets.
Next, through comprehensive experiments, we systematically~compare state-of-the-art retrosynthesis prediction models on the two groups of benchmarks, revealing the limitations of previous in-distribution evaluation and re-examining the advantages of each model. 
More remarkably, we are motivated by the above empirical insights to propose two model-agnostic techniques that can improve the OOD generalization of arbitrary off-the-shelf retrosynthesis prediction algorithms. 
Our preliminary experiments show their high potential with an average performance improvement of $4.6\%$, and the established benchmarks serve as a foothold for further retrosynthesis prediction research towards OOD generalization.
\end{abstract}

\section{Introduction}

Retrosynthesis is the fundamental step in the field of organic synthesis~\cite{corey}, which involves the application of various strategies to break down a target molecule into simpler building-block molecules. One of the biggest challenges for the pharmaceutical industry is finding reliable and effective ways to make new compounds. 

Recently, there has been growing interest in computer-aided synthesis planning due to its potential to reduce the effort required for manually designing retrosynthesis strategies with chemical knowledge. Numerous machine learning models have been developed to learn these strategies from a fixed training dataset and to generalize this knowledge to new molecules. The training process involves either learning an explicit set of hard-coded templates as fixed rules in a template-based manner or learning an implicit high-dimensional mapping from the product to the precursors in a template-free approach. Under the standard independent and identically distributed~(i.i.d.) train-test data split, both lines of approaches have yielded promising results.

\begin{figure*}[htb]
\begin{center}
\centerline{\includegraphics[width=0.9\linewidth]{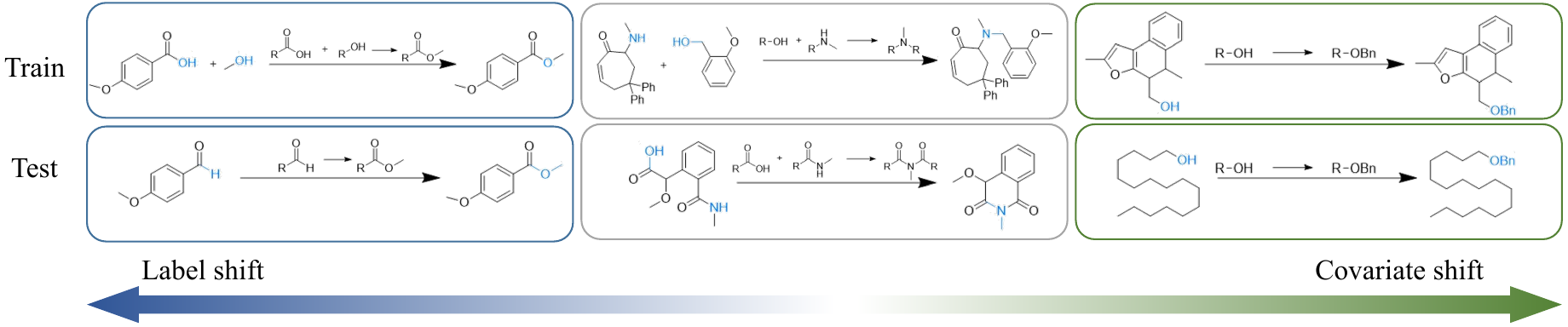}}
\caption{Three train-test pairs of reactions exhibiting label shift OOD, regular ID, and covariate shift OOD (from left to right) were analyzed. In the case of label shift OOD, a distributional shift manifests in retro-strategies within similar target molecules. Conversely, in the case of covariate shift OOD, a distributional shift occurs in target molecules within similar retro-strategies.}
\label{fig:intro}
\end{center}
\end{figure*}

However, all retrosynthesis models exhibit a performance discrepancy between an in-distribution~(ID) and an out-of-distribution~(OOD) test set, which is a common issue when deploying the retrosynthesis model to a real-world environment. In general, we conclude that this discrepancy can be essentially attributed to the two different types of distributional shifts: the \textbf{label shift} of the retrosynthesis strategies~(retro-strategy) and the \textbf{covariate shift} of the target molecules. Our research is aimed to analyze this discrepancy caused by the distributional shifts under various retrosynthesis prediction baselines. By understanding the distributional shifts that occur, we hope to mitigate their effects and gain a deeper understanding of the behaviors of various types of retrosynthesis models under different distributional shifts. To the best of our knowledge, no study has been conducted to investigate this topic rigorously.

To systematically analyze these two distributional shifts, we create and conduct experiments on multi-dimensional OOD dataset splits using various types of single-step retrosynthesis prediction baseline models. For each type of distributional shift, we propose a model-agnostic approach to alleviate the performance degradation. Our paper contributes to this field in three ways:
\begin{itemize}
    \item We systematically define the distributional shifts under the context of retrosynthesis prediction.
    \item We construct multi-dimensional out-of-distribution datasets benchmarks and analyze the degree of performance discrepancies on a comprehensive set of baselines.
    \item We propose model-agnostic invariant learning and concept enhancement techniques to reduce performance degradations and provide our insights.
\end{itemize}

\section{Related Work}

 \textbf{Single-step retrosynthesis prediction} The role of machine learning in retrosynthesis prediction is becoming increasingly pronounced~\cite{JIANG202332}, especially in the pivotal stage of single-step \retro prediction.
 Single-step \retro aims to predict the set of molecules that chemically react to form a given product, towards which existing approaches fall into three major categories, including template-based~(TB), semi-template-based~(semi-TB), and template-free~(TF) ones. 
 Templates~\cite{template} encode the changes in atom connectivity during the reaction, thereby applicable in converting a product back into the corresponding precursors. 
 TB approaches such as NeuralSym~\cite{neuralsym}, Retrosim~\cite{retrosim}, and GLN~\cite{gln} learn to select a standard
 reaction template to apply to the specified product for deriving the resulting precursors with subgraph isomorphism.
 However, TB methods have been criticized for their poor generalization capability to reactions outside the underlying training template set~\cite{tb1, neuralsym, wln}. 
 Semi-TB models alleviate the generalization problem via either constructing a more flexible template 
 database with subgraph extraction~\cite{localretro,retrocomposer} or decomposing \retro into two sub-tasks of i) center identification and ii) synthon completion~\cite{retroxpert,graphretro}.
On the other hand, TF approaches completely eliminate using reaction templates and instead learn chemical transformations implicitly. Using various molecule representations, existing TF solutions formulate \retro as a string~\cite{mt0, mt, dual-tf,yu2022grasp} or a graph~\cite{g2gs,megan} translation problem.


\textbf{OOD generalization for molecule-related tasks } 
Notwithstanding extensive literature on the evaluation of ID generalization, some attempts have been made to explore the frequent distributional shifts in real-world molecule-related tasks, including retrosynthesis prediction.
The works~\cite{drugood,OOD2,OOD3} systematically study 
the shift in molecular size and structure as well as labels, and present several OOD benchmark datasets. 
However, they set their sights on molecular property prediction for drug discovery, which substantially differs from \retro prediction as a molecular generation task. 
Molecular generation also introduces additional complexity in the definition of label space, which is more complicated than a single value in conventional regression and classification in property prediction.
As a matter of fact, there have been some works devoted to investigating the factors that cause label shift in \retro prediction, including the change in template radius, size, and subgraph isomorphism~\cite{temp1,temp2,temp4} between training and testing reactions.
Unfortunately, the influence of such label shift on the performance of existing single-step \retro prediction approaches remains largely unknown, though the two approaches in~\cite{mhn} and 
\cite{unseen} as a TB and TF approach respectively attempt to evaluate the ``zero-shot'' reaction prediction performances.
However, the definitions of ``zero-shot" used are arbitrary and lack consistency, with \cite{mhn} considering new reaction templates as "zero-shot", while \cite{unseen} defines "zero-shot" samples as new reaction types.
Besides the lack of comprehensive performance evaluation under label shift, benchmark datasets that support such evaluation are also in urgent demand.
Existing dataset splits for distribution shift in \retro prediction, either by reaction type bias~\cite{temp5} or by time period~\cite{segler2018}, struggle to explicitly disentangle label shift from covariate shift. 
These early exploratory studies motivate a more rigorous and systematic analysis of the impact of distribution shift on \retro prediction, covering (1) the disentanglement of two types of shift, (2) benchmark datasets for each type of shift, (3) extensive empirical evaluation of state-of-the-art \retro prediction algorithms, and (4) two model-agnostic techniques to handle both shifts.

\section{Preliminaries}

In this section, we formally define the distributional shift in single-step retrosynthesis prediction and establish the notation used throughout the paper. 

\subsection{Out-Of-Distribution Retrosynthesis Prediction}

Single-step \retro prediction is a task where the model receives a target molecule $m \in \mathcal{M}$ as input and predicts a set of precursor source precursors $r \in \mathcal{R}$ that can synthesize $m$. The model can use different molecular transformation rules to generate various precursors for target molecules. Depending on the definitions of these transformation rules, retrosynthesis models can be classified into two main categories: \textit{template-based} and \textit{template-free}. Template-based approaches utilize reaction templates to predict the precursors required for synthesizing a product. These templates encode the changes in atom connectivity during the reaction that represent a specific type of molecular transformation. On the other hand, template-free models use a generative model to generate the precursors for a given target directly. These models typically use the SMILES~\cite{smiles} string or a graph structure to represent molecules and implicitly learn high-dimensional transformation rules between the hidden representations of precursors and molecules. However, such transformation rules can always be mapped back to reaction templates after the reaction generation.

Without loss of generality, we denote the retro-strategy as $t \in \mathcal{T}$ to represent such transformation rules from target molecule $m \in \mathcal{M}$ to source precursors $r \in \mathcal{R}$, which are meant to be general and not specific to any particular model or approach. $\mathcal{T}$ 
 represents the space of transformation rules applied to a target molecule to generate its precursors. It's crucial to note that our introduction of $\mathcal{T}$ is not merely restricted to a template-based interpretation. In essence, all retrosynthesis prediction models, in an end-to-end fashion, intake a target product and output a set of precursors. While the exact realization of these retro-strategies might differ among models, our evaluation still remains model-agnostic and is conducted solely on the exact matching of the output precursors.


Subsequently, the training and testing datasets for retrosynthesis are denoted as $\mathcal{D}^{tr}=\{(m_i, t_i)\}_{i=1}^{N}$ and $\mathcal{D}^{tst}=\{(m_i^*, t_i^*)\}_{i=N+1}^{N+N^{*}}$. 
The out-of-distribution \retro prediction problem can be defined as follows:

\begin{definition}
Given the observational training reactions $\mathcal{D}^{tr}=\{(m_i, t_i) \sim P_{tr}(\mathcal{M},\mathcal{T})\}_{i=1}^{N}$ and testing data $\mathcal{D}^{tst}=\{(m_i^{*}, t_i^{*}) \sim P_{tst}(\mathcal{M},\mathcal{T})\}_{i=1}^{N^{*}}$ where $ P_{tr}(\mathcal{M},\mathcal{T}) \neq  P_{tst}(\mathcal{M},\mathcal{T})$ and  $N/N^{*}$ is the sample sizes of train/test data, the goal of out-of-distribution retrosynthesis prediction is to learn a model in training distribution $P_{tr}(\mathcal{M},\mathcal{T})$ to generalize to the test distribution $P_{tst}(\mathcal{M},\mathcal{T})$ accurately.
\end{definition}

\section{OOD Retrosynthesis Prediction Benchmarks}

In this section, we rigorously define and investigate two types of distributional shifts in the context of retrosynthesis: \textbf{label shift} of retro-strategies, $P(\mathcal{T})$, and \textbf{covariate shift} of target molecules, $P(\mathcal{M})$. Subsequently, we create two out-of-distribution dataset splits for each shift on the benchmark retrosynthesis prediction dataset under different domain settings. These datasets are used in subsequent empirical studies to analyze performance gaps and evaluate the effectiveness of our proposed OOD generalization approaches.

\begin{figure*}[htb]
\begin{center}
\centerline{\includegraphics[width=0.85\linewidth]{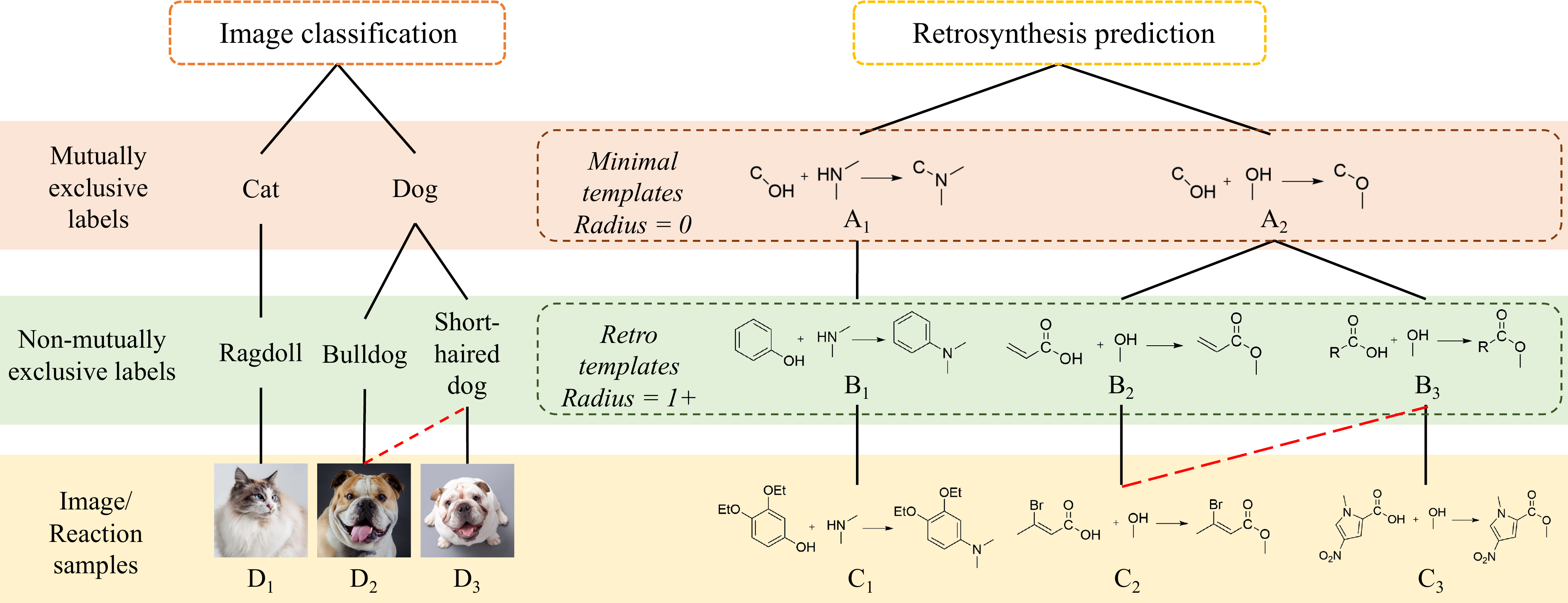}}
\caption{Minimal-templates and retro-templates. \textbf{Left}: In the image classification task, cats and dogs are typically regarded as mutually exclusive, while ragdolls, bulldogs, and short-haired dogs are not. Though image $D_2$ is only annotated as a bulldog, it has the potential label of a short-haired dog. \textbf{Right}: In retrosynthesis, similar to the image $D_2$, both labels $B_2$ and $B_3$ are viable options for generating the correct reaction $C_2$, but only one template $B_2$ is exposed as the positive label in the training dataset due to non-mutual exclusivity.}
\label{fig:tpl_illustration}
\end{center}
\end{figure*}

\subsection{Label Shift in Retro-strategy $P(\mathcal{T})$}

In the context of retrosynthesis prediction, we define the label space as the set of retro-strategies, denoted as $\mathcal{T}$, that map from the space of target molecules, $\mathcal{M}$, to the space of precursors, $\mathcal{R}$. In general, the label shift refers to the change of distribution of retro-strategy $P_{tr}(\mathcal{T}) \neq P_{tst}(\mathcal{T})$. However, the definition of the retro-strategy can vary significantly among different types of retrosynthesis prediction models. For template-based models, the retro-strategy is a discrete set of reaction templates extracted from the training set during data pre-processing. On the other hand, for template-free models, the retro-strategy is learned inherently during training, which is a function space that maps $\mathcal{M}$ to $\mathcal{R}$ in the latent space. It is widely acknowledged in studies~\cite{temp2, temp1, lin2020auto, mt} that template-free models can generalize to novel or unseen reaction templates, whereas template-based models are confined to the predefined set of extracted templates. 

Nevertheless, our findings reveal that the claimed generalization ability of the template-free models highly depends on the granularity of templates. As shown in Fig.~\ref{fig:tpl_illustration}, we focus on two different granularity of templates: minimal-template~(radius=0) and retro-template~(radius=1+). The key difference is that a reaction can \textit{only} be mapped to one distinct minimal template, while it is possible to be mapped to multiple retro-templates. Although almost all previous template-based methods used retro-template as the template definition, we discover that the nuisance in retro-strategy granularity will result in distinct performance differences in the OOD label shift. We provide a more detailed investigation of template granularity in the Appendix.


\begin{figure}[htb]
\begin{center}
\centerline{\includegraphics[width=\columnwidth]{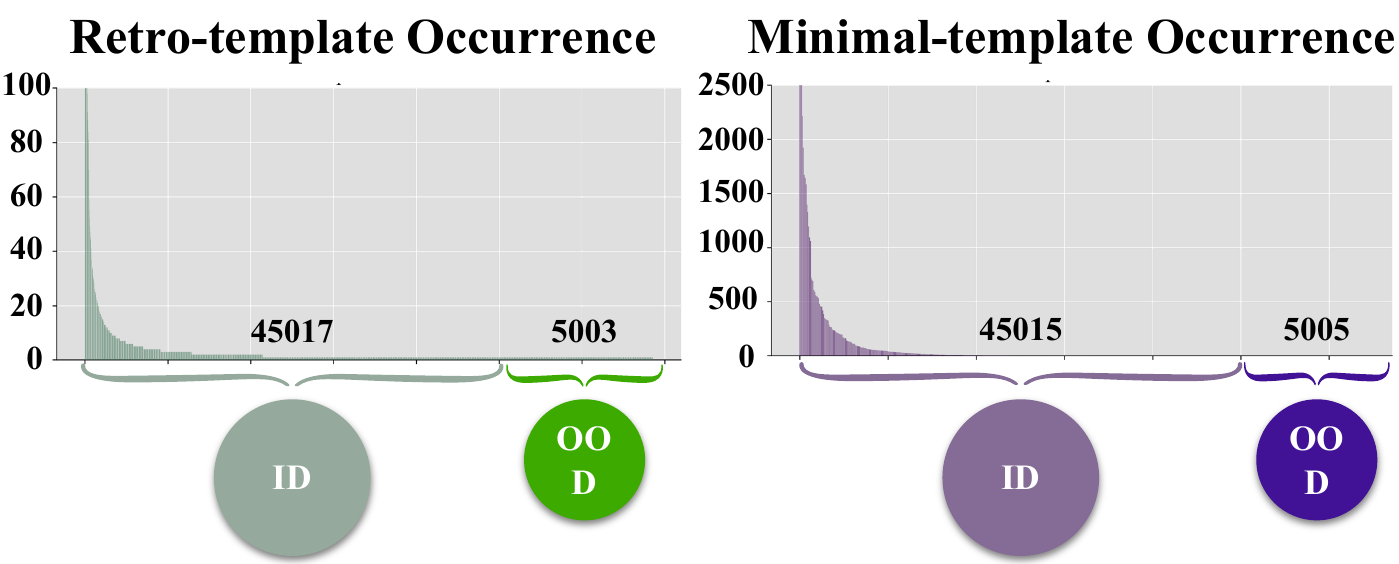}}
\caption{Retro-template and minimal-template distribution on USPTO50K and the ID/OOD dataset split for label shift dataset \texttt{USPTO50K\_T}.}
\label{fig:tpl_count}
\end{center}
\end{figure}

\begin{figure}[htb]
\begin{center}
\centerline{\includegraphics[width=\columnwidth]{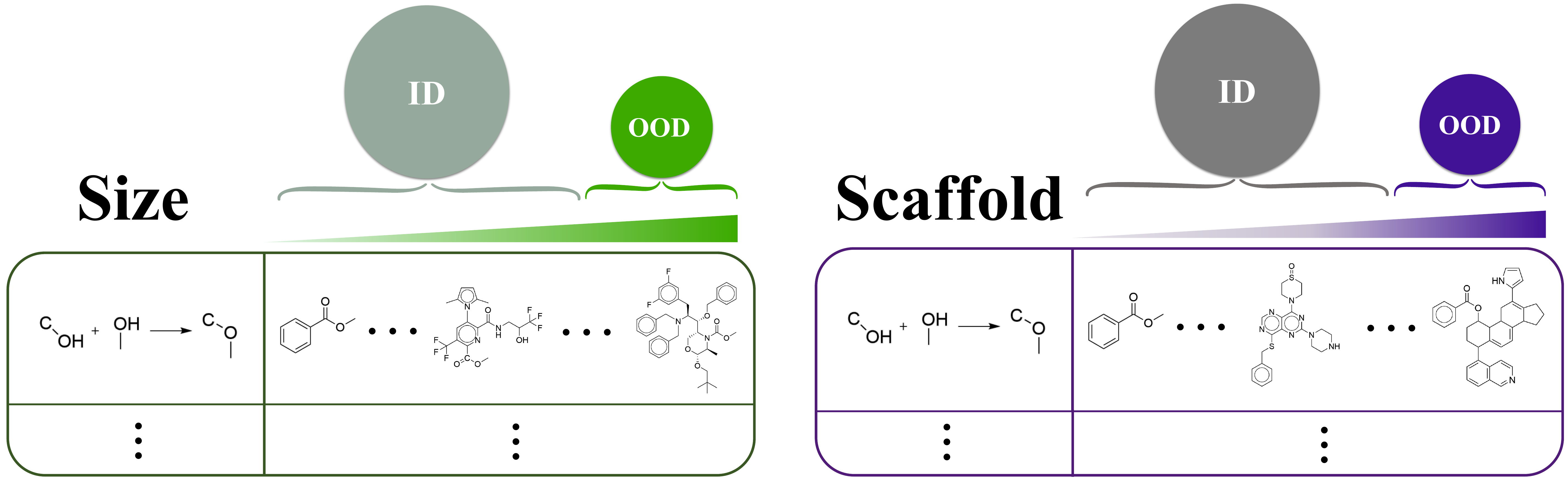}}
\caption{Molecular size and scaffold ID/OOD dataset split for covariate shift dataset 
\texttt{USPTO50K\_M}.}
\label{fig:cov-shift}
\end{center}
\end{figure}


\subsection{Covariate Shift in Target Molecule $P(\mathcal{M})$}

In the context of retrosynthesis prediction, covariate shift refers to the change in the distribution of the target molecules as $P_{tr}(\mathcal{M}) \neq P_{tst}(\mathcal{M})$. This phenomenon is often studied in conjunction with the concept distribution $P(\mathcal{T}|\mathcal{M})$, as the fundamental assumption for accurately evaluating covariate shift is that the concept distribution remains constant, $P_{tr}(\mathcal{T}|\mathcal{M}) = P_{tst}(\mathcal{T}|\mathcal{M})$. Typically, previous works\cite{inv1, irmv1} addressed the covariate shift by adopting a causal perspective and dividing the input into two separate parts: the invariant feature $\mathcal{M}\_{inv}$ and the variant~(spurious) feature $\mathcal{M}\_{var}$. The invariance property holds that using the invariant feature $\mathcal{M}\_{inv}$ alone is sufficient to fully recover the concept, such that $P(\mathcal{T}|\mathcal{M})=P(\mathcal{T}|\mathcal{M}\_{inv})$. Therefore, a pure covariate shift dataset should be designed in such a way that all shifts in the distribution occur on the variant feature $\mathcal{M}_{var}$ when $P_{tr}(\mathcal{M}) \neq P_{tst}(\mathcal{M})$ to maintain the invariance properties.

Covariate shift in molecular structures is prevalent in molecular property prediction and material design tasks. Similarly, the invariance property assumes that specific patterns inside a molecule, such as functional groups or scaffold substructures, play a crucial role in predicting a specific property. Generally, the substructure invariance rules that govern these relationships are task-specific for each property and have been validated through extensive study and observation as prior knowledge.~\cite{subinvariant1,subinvariant3,subinvariant2} 
Following the same concept, we assume that certain features or substructures $\mathcal{M}\_{inv}$ in the target molecule are crucial for the model to make an invariant prediction for different retro-strategies to maintain $P(\mathcal{T}|\mathcal{M})=P(\mathcal{T}|\mathcal{M}\_{inv})$. Naturally, the reaction center~(radius=0) should always be included as part of the invariant feature; otherwise, applying the templates to the target molecule would result in automatic failure. In addition, other substructures not limited to the reaction center can simultaneously impact the applicability of a particular template in terms of chemo-, regio-, or stereo-selectivity, which are the features we aim to identify as additional parts of $\mathcal{M}_{inv}$. 

\subsubsection{OOD Benchmark Dataset Split}

 We introduce the benchmark dataset construction process for label shift and covariate shift dataset split. The detailed construction process on the benchmark is elaborated in the Appendix.
 
 \textbf{Label shift benchmark dataset }To systematically evaluate the generalizability of different models when facing label shift in retro-strategies, we generate two OOD dataset splits as \texttt{USPTO-50K\_T}, on the benchmark USPTO50K dataset~\cite{uspto50k} using the different granularity of labels. As shown in Fig.~\ref{fig:tpl_count}, we extract the minimal-templates and retro-templates for each reaction and arrange them in descending order based on their occurrence frequency. We also deliberately ensured that the total template set does \textit{not} intersect between ID and OOD subsets to investigate the ability of different models to generalize to novel retro-strategies.

 \textbf{Covariate shift benchmark dataset }We adopt the similar definition of covariate split settings proposed in ~\cite{wilds,drugood} by using the molecular size and scaffold as criteria to construct the covariate OOD dataset split \texttt{USPTO-50k\_M}. The approach involves arranging the samples based on the molecular size or scaffold differences of the target molecule in ascending order and selecting larger or more complex target molecules as the OOD subset.  Generally, a target molecule with a larger size or a more complex scaffold contains a larger proportion of variant features $\mathcal{M}\_{var}$~\cite{wilds,drugood}. As shown in Fig.~\ref{fig:cov-shift}, to eliminate the irrelevant influence of label shift, we conduct the data split independently for \textit{each} minimal-template and then combine the results. This approach ensures that all covariate shifts occur on the variant feature $\mathcal{M}_{var}$ during $P_{tr}(\mathcal{M}) \neq P_{tst}(\mathcal{M})$, and guarantees that the ground-truth disconnection site stays consistent among all samples within a specific template class.



\section{State-of-the-art Retrosynthesis Prediction Models under Distributional Shift}

In this section, we introduce five representative retrosynthesis prediction models for empirical studies and analyze their baseline performance under the two distributional shifts mentioned above. 

\subsection{Baseline Methods} 

We select five representative models, namely \textit{GLN}~\cite{gln}, \textit{Molecular Transformer~(MT)}~\cite{mt}, \textit{GraphRetro~(G\_Retro)}~\cite{graphretro}, \textit{RetroComposer~(R\_Composer)}~\cite{retrocomposer}, and \textit{MHN}~\cite{mhn},  as our baseline methods for empirical studies. These models comprehensively cover SMILES-based and graph-based representation under template-based, semi-template-based, and template-free categories and the detailed introduction of the baseline methods can be found in the Appendix. All baseline models are re-trained on each of the four OOD datasets separately for evaluation by substituting the original USPTO50K split with the respective OOD split. We use the widely accepted top-k accuracy as the evaluation metric, which is still the most appropriate quantifiable metric for retrosynthesis prediction.

\begin{table*}[ht]
\begin{center}
\begin{small}
\begin{sc}
\resizebox{0.9\linewidth}{!}{
\begin{tabular}{l|ccccccccccr}
\toprule
mol-size  & GLN\_base & GLN\_irm  &MT\_base & MT\_irm & G\_Retro\_base & G\_Retro\_irm 
& R\_Composer\_base & R\_Composer\_irm & MHN\_base & MHN\_irm\\
\midrule
ID Top-1   & 54.5\% & 54.9\% & 52.5\% & 52.2\% & 54.9\%  & \textbf{55.8\%}  & 55.1\% & 55.3\% & 52.5\%  & 52.2\%\\
ID Top-3   & 69.0\% & 70.5\%  & 75.7\% & 76.1\% & 70.7\% &  71.1\% & 79.7\% & \textbf{80.5\%} & 76.4\%  & 76.6\%\\
ID Top-5  & 76.9\% & 77.6\% & 80.1\% & 81.4\% & 74.7\%  &  75.4\% & 86.0\% & \textbf{87.7\%} & 84.0\%  & 84.2\%\\
ID Top-10  & 85.1\% & 85.7\% & 83.6\% & 85.0\% & 77.7\% &  78.2\% & 90.0\% & \textbf{90.9\%} & 89.8\%  & 90.2\%\\
\midrule
OOD Top-1   & 37.6\% & 38.0\% & 29.9\%  & 30.3\% & 38.5\% &  39.6\% & 41.2\% & \textbf{41.6\%} & 34.0\%  & 33.8\%\\
OOD Top-3   & 50.7\% & 51.2\% & 46.7\% & 47.7\% & 57.2\% &  58.4\% & 67.3\% & \textbf{68.2\%} & 57.3\%  & 57.5\%\\
OOD Top-5  &  58.9\% & 59.4\% & 53.8\% & 55.0\%  & 64.5\% &   65.9\% & 75.1\% & \textbf{76.3\%} & 67.5\%  & 67.9\%\\
OOD Top-10  & 70.7\% & 71.5\% & 58.1\% & 60.1\%  & 70.9\% &  72.8\% & 83.1\% & \textbf{84.9\%} & 78.9\%  & 79.2\%\\
\bottomrule
\toprule

mol-scaffold  & GLN\_base & GLN\_irm  &MT\_base & MT\_irm & G\_Retro\_base & G\_Retro\_irm 
& R\_Composer\_base & R\_Composer\_irm & MHN\_base & MHN\_irm\\
\midrule
\midrule
ID Top-1   & 55.7\% &  56.0\% & 51.8\% & 51.4\% & 56.2\% & \textbf{56.2\%} & 52.4\% & 52.8\% & 52.3\%  & 51.9\% \\
ID Top-3   & 69.8\% & 70.7\% &  75.5\% & 76.0\% &70.2\% & 71.2\% & 78.2\% & \textbf{79.2\%} & 76.6\%  & 76.5\% \\
ID Top-5  & 77.2\% & 77.9\% & 80.3\% & 81.6\%  &73.8\% & 74.6\%  & 84.8\% & \textbf{86.4\%} & 84.0\%  & 84.3\%\\
ID Top-10  & 85.5\% & 86.1\% & 82.9\% & 84.7\%  &76.6\%  & 77.1\% & 89.5\% & \textbf{90.3\%} & 90.1\%  & 90.2\%\\
\midrule
OOD Top-1   & 38.9\% & 39.5\% & 37.9\% & 38.6\% &39.9\% & 40.1\%  & 40.7\% & \textbf{41.2\%} & 35.1\%  & 34.8\%\\
OOD Top-3   & 53.3\% & 53.9\% & 57.7\% & 59.0\% &57.8\% & 59.7\%  & 65.5\% & \textbf{66.4\%} & 60.3\%  & 60.4\%\\
OOD Top-5  &  61.0\% & 61.7\% & 64.0\% & 65.6\% &64.7\% & 66.5\%    & 75.1\% & \textbf{76.2\%} & 69.4\%  & 69.7\%\\
OOD Top-10  & 72.4\% & 73.5\% & 69.2\% & 70.2\% & 71.2\% & 73.4\%  & 82.6\% & \textbf{84.5\%} & 79.9\%  & 80.1\%\\
\bottomrule
\end{tabular}
}
\end{sc}
\end{small}
\end{center}
\caption{The performance of five baselines and their IRM variants on covariate shift $P(\mathcal{M})$. The best IRM result is reported with \textit{center-token masking IRM} for MT, \textit{center prediction IRM} for GLN, \textit{graph edit IRM} for GraphRetro, and \textit{template composer IRM} for R\_Composer, respectively.}
\label{res:covariate}
\end{table*}

\subsection{Baseline Results}

The baseline results under covariate shift and label shift are listed in Tab.~\ref{res:covariate} and Tab.~\ref{res:label}, respectively, with subscript $_{base}$. 

\textbf{Covariate shift} In Tab.~\ref{res:covariate}, we note a significant decline in performance, specifically a reduction of 30-40\% in top-1 accuracy when comparing the ID test set to the OOD set. The present findings support our prior hypothesis that larger molecules introduce more complexity in predicting the correct retro-strategy, regardless of the model employed, and these complexities are limited to specific feasible disconnection sites for a given target molecule. Among the five baseline models, MT exhibits the most significant decline in performance, since larger target molecules result in larger precursors with longer sequences of SMILES tokens as error accumulation, thereby intensifying the challenges associated with the covariate shift. RetroComposer outperforms most baselines in both splits, which can be attributed to its subgraph selection mechanism in discovering robust substructure invariance within the training samples.

\textbf{Label shift} In Tab.~\ref{res:label}, the results are more varied between retro-template and minimal-template split. The average performance degradation is around 40-50\% in the retro-template split and almost 100\% in the minimal-template split. For retro-template split, we conclude that it's not rigorous to assume that template-based approaches cannot generalize to new templates without specifying the radius boundary, since our result shows that both GLN and MHN successfully generalize to a portion of unseen retro-templates due to their non-mutually exclusive nature. Additionally, we discover that when facing the same label shift in retro-templates, template-free models do not exhibit a clear advantage over template-based models in generalizability. On the other hand, the results in minimal-template split align with the previously held assumption that template-free models have only a limited ability to generalize to unseen templates when compared with template-based models.

\begin{table*}[ht]
\begin{center}
\begin{small}
\begin{sc}
\resizebox{0.9\linewidth}{!}{
\begin{tabular}{l|ccccccccccr}
\toprule
Retro-template  & GLN\_base & GLN\_irm  &MT\_base & MT\_irm & G\_Retro\_base & G\_Retro\_irm 
& R\_Composer\_base & R\_Composer\_irm & MHN\_base & MHN\_irm\\
\midrule
ID Top-1    & 50.9\% & 51.8\% & 47.1\% & 49.0\% & 53.2\%  & \textbf{53.9\%} & 53.0\% & 53.3\% & 51.9\% &	52.2\% \\
ID Top-3    & 68.2\% & 70.3\% & 64.6\% & 67.0\% & 68.6\% &  69.4\% & 78.1\% & \textbf{78.5\%} & 75.1\% & 75.8\% \\
ID Top-5   & 76.2\% & 78.4\% & 68.1\% & 72.9\% & 72.3\%  &  73.5\% & 85.2\% & \textbf{86.4\%} & 82.7\% & 83.4\% \\
ID Top-10   & 85.2\% & 87.8\% & 71.2\% & 77.1\% & 74.8\% &  75.7\% & 90.2\% & \textbf{90.9\%} & 89.9\% & 90.7\% \\
\midrule
OOD Top-1   & 22.9\% & 24.5\% & 23.8\% & 25.2\% & 27.0\% & \textbf{28.6\%}  & 25.4\% & 26.9\% & 18.7\% &	20.4\% \\
OOD Top-3   &  31.8\% & 36.6\% & 35.8\% & 41.2\% & 40.3\%  & 42.5\%  & 41.7\% & \textbf{43.5\%} & 33.1\% &	36.1\% \\
OOD Top-5   & 38.8\% & 43.4\% & 39.8\% & 48.7\% & 44.3\%  & 46.6\%  & 	47.6\% & \textbf{49.8\%} & 40.5\% &	42.8\% \\
OOD Top-10   & 46.6\% & 52.6\% & 43.9\% & \textbf{55.9\%} & 47.4\%  & 49.4\%  & 52.9\% & 55.4\% & 49.6\% &	52.4\% \\
\bottomrule
\toprule

Minimal-template  & GLN\_base & GLN\_irm  &MT\_base & MT\_irm & G\_Retro\_base & G\_Retro\_irm 
& R\_Composer\_base & R\_Composer\_irm & MHN\_base & MHN\_irm\\
\midrule
\midrule
ID Top-1    & 51.9\% & 53.3\% & 48.1\% & 49.5\% & 53.6\%  & 54.2\% & 53.9\% & \textbf{54.2\%} & 52.9\% &	53.1\% \\
ID Top-3    & 68.9\% & 69.9\% & 64.6\% & 67.2\% & 68.3\%  & 69.9\% & 78.6\% & \textbf{79.3\%} & 74.2\% &	74.7\%  \\
ID Top-5   & 76.5\% & 78.3\% & 69.9\% & 73.3\% & 74.8\%  & 76.4\% & 85.6\% & \textbf{86.7\%} & 83.2\% &	83.9\%  \\
ID Top-10  & 86.6\% & 88.9\% & 74.2\% & 80.2\% & 76.4\%  & 79.1\% & 89.7\% & 90.5\% & 90.6\% &	\textbf{91.3\%}  \\
\midrule
OOD Top-1    & 0\% & 0\% & \textbf{2.8\%} & 2.3\% & 0\%  & 0\% & 0.1\% & 0.1\% & 0.0\% &	0.1\%  \\
OOD Top-3    & 0\% & 0\% & 3.8\% & \textbf{4\%} & 0.1\%  & 0.2\% & 0.4\% & 0.4\% & 0.1\% &	0.1\%  \\
OOD Top-5   & 0\% & 0.2\% & 4.2\% & \textbf{4.7\%} & 0.1\%  & 0.3\% & 0.7\% & 0.9\% & 0.2\% &	0.2\%  \\
OOD Top-10   & 0\% & 0.2\% & 5.0\% & \textbf{5.7\%} & 0.1\%  & 0.3\% & 1.2\% & 1.2\% & 0.3\% &	0.4\%  \\
\bottomrule
\end{tabular}
}
\end{sc}
\end{small}
\end{center}
\caption{The performance of five baselines and enhanced versions on label shift $P(T)$. The best-enhanced result is reported with $n=5$ for MT,GLN, and MHN, and $n=2$ for GraphRetro and Retrocomposer.}
\label{res:label}
\end{table*}

\section{Invariant Learning for Covariate Shift}

In order to handle the covariate shift, our objective is to learn a robust parametric representation $\Phi(\cdot)$ that can accurately capture the full invariant features in the target molecule that satisfies the invariance property for predicting the retro-strategy. Specifically, we adopt Invariant Risk Minimization~(IRM) to learn this invariant feature representation, which requires that the feature representation is simultaneously optimal across different domains. While the IRM regularizer is a known model-agnostic method, its precise application and optimization in a retrosynthesis model is a non-trivial problem and needs to be handled carefully in a model-specific way. We elaborate on the detailed IRM implementation for each baseline in the Appendix.


\subsection{Performance Analysis}

The best results are listed in Table~\ref{res:covariate} for the \texttt{USPTO50k\_M}. Overall, we discover that applying IRM regularization to the specific reaction center identification stage improves the performance of GLN, GraphRetro, and RetroComposer, but the improvement is marginal for MT and MHN due to the nature of sequence-to-sequence generation and entanglement of center prediction and precursor generation, respectively. The overall insignificant improvement using IRM can be attributed to the uncontrollable concept drift on $P(\mathcal{T}|\mathcal{M}_{inv})$ presented within the dataset. The reason is that the collection of the reactions in USPTO50K is subject to the prior selection bias of different chemists during distinct wet-lab experiments under unobserved covariates~(such as chemical conditions, etc.). Therefore, the ideal assumption for the invariance property is often violated, resulting in incoherency from the concept drift that hinders IRM from learning an optimal invariant predictor. In addition, the improvement resulting from the application of IRM could be more substantial if the distribution of training data were less biased towards specific retro-strategies.


Apart from the statistical result, we observe that using IRM regularization reduces spurious correlation on variant substructures $\mathcal{M}_{var}$ and an increased convergence towards the invariant substructures $\mathcal{M}_{inv}$ as shown in Fig.~\ref{fig:visualization_analysis}. We also comprehensively evaluated the results of applying IRM regularizer to different loss components as an ablation study presented in Tab.~\ref{res:cov-ablation} in the Appendix.

\begin{figure*}[ht]
\begin{center}
\captionsetup{width=0.9\linewidth}
\centerline{\includegraphics[width=0.8\linewidth]{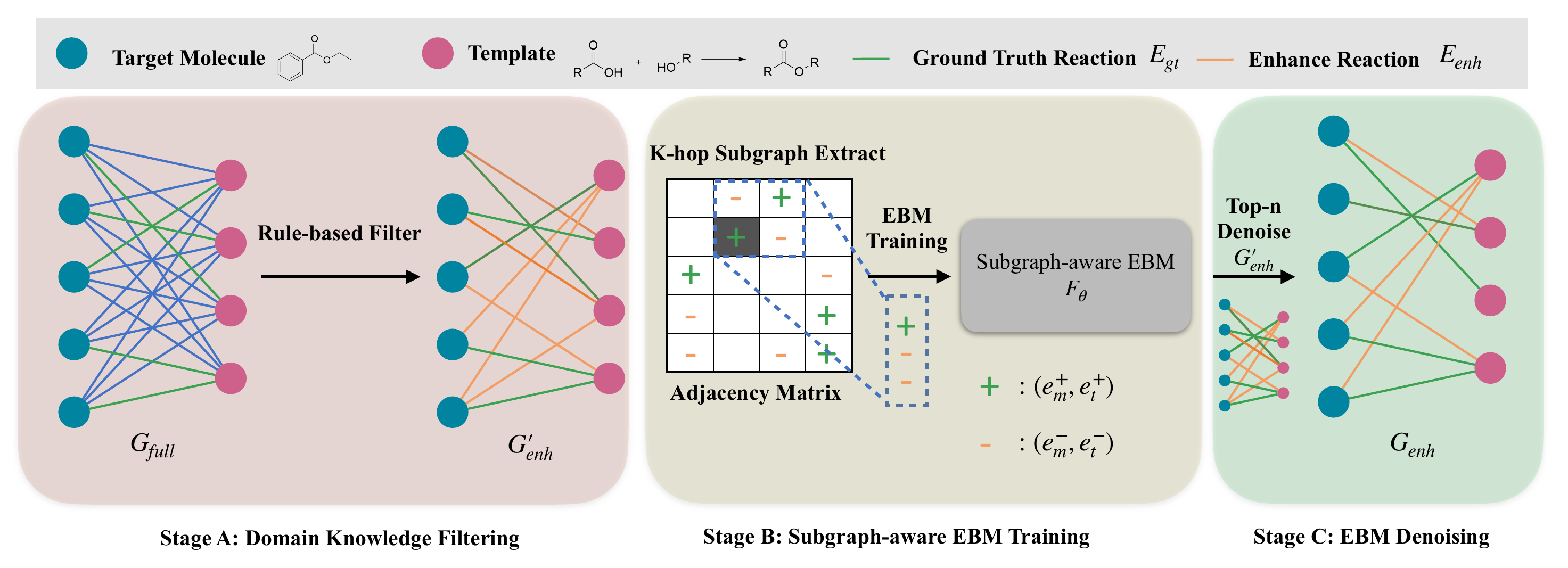}}
\caption{The process of acquiring the enhanced bipartite graph $G_{enh}$ between target molecules and templates. Stage. A: We use a rule-based approach utilizing domain knowledge to filter $G_{full}$ to obtain a potentially enhanced graph $G_{enh}^{\prime}$. Stage B: We utilize the k-hop reaction-level subgraph extraction method to construct the subgraph-aware tractable loss to train the EBM model. The demonstration in the figure illustrates the 1-hop subgraph extraction under the edge with the grey background in the adjacency matrix. Stage C: We use the trained EBM to denoise the potentially enhanced graph $G_{enh}^{\prime}$ and obtain the final enhanced graph $G_{enh}$. }
\label{fig:demo}
\end{center}
\end{figure*}

\section{Concept Enhancement for Label Shift}

Besides covariate shift in molecular space, retrosynthesis prediction suffers significantly from the label shift $P(\mathcal{T})$. The reason behind the label shift is that the current benchmark dataset only includes reactions deemed most favorable by different chemists, thus manifesting a high precision. Still, it indiscriminately regards other unobserved potentially feasible reactions as equally infeasible, resulting in a low recall. Essentially, retrosynthesis is a many-to-many problem~\cite{thakkar2022unbiasing, schwaller2019evaluation}, where the target molecule $M$ can potentially be synthesized through various distinct retro-strategies $T$, and vice versa. To mitigate the low recall issue, we aim to enhance the concept of template applicability by transforming the binary criteria of the observed ground truth $P_{gt}(\mathcal{M}, \mathcal{T})$ into a continuous approximation using a probabilistic model. By utilizing a probabilistic model, we have greater flexibility to evaluate the boundary cases from the potentially feasible reactions, thereby constructing a more robust training set and improving recall without compromising precision.

 However, modeling such probability is non-trivial since we need to perform counterfactual inference of unobserved reactions. One intuitive approach is to assume the distribution $P(\mathcal{M}, \mathcal{T})$ follows a Gaussian Process~(GP) in order to construct a posterior predictive distribution for the feasibility of unobserved reactions. However, this assumption has limited expressiveness and may over-simplify the complex probabilistic structure of the selection bias among chemists. 
 
 Inspired by the recent advancements of Energy-based Model~(EBM)~\cite{ebm1, ebm2020}, which offers greater flexibility and expressiveness compared to traditional probabilistic models, we adopt the EBM architecture to approximate $P(\mathcal{M}, \mathcal{T})$. EBM represents the likelihood of a probability distribution $p_{D}(\mathbf{x})$ for $x \in \mathbb{R}^{D}$ as $p_{\theta}(\mathbf{x}) = \frac{\exp(-F_{\theta}(\mathbf{x}))}{Z(\theta)}$, where the function $F_{\theta}(\mathbf{x}): \mathbb{R}^{D} \rightarrow \mathbb{R}$ is known as the energy function, and $Z(\theta)=\int_{\mathbf{x}}\exp(-F_{\theta}(\mathbf{x}))$ is known as the partition function. Typically, directly evaluating $p_{\theta}(\mathbf{x})$ requires an intractable integration in partition function $Z(\theta)$ over all possible target-template tuples.  Fortunately, the gradient for training the EBM, $\nabla_{\theta} \log p_{\theta}(\mathbf{x})$, can be expressed in the alternative form:

 \begin{small}   
 \begin{equation}
     \nabla_\theta \log p_\theta(\mathbf{x})=\mathbb{E}_{p_\theta\left(\mathbf{x}^{\prime}\right)}\left[\nabla_\theta F_\theta\left(\mathbf{x}^{\prime}\right)\right]-\nabla_\theta F_\theta(\mathbf{x})
    \label{eq:gradient}
 \end{equation}
\end{small}
\par
 Thus, the question left for us is finding a surrogate for samples $\mathbf{x^{\prime}}$ from the distribution $p_(\theta)(\mathbf{x^{\prime}})$ to approximate the gradient of the training loss. In the next section, we elaborate on using a k-hop subgraph extraction algorithm on a bipartite graph to build the tractable EBM training loss. 
 
\subsection{Approach}

 The complete enhancement process is shown in Fig.~\ref{fig:demo}. To begin, we model the set of ground-truth reactions as a target-template bipartite graph, where the ground-truth graph $G_{gt} = (M, T, E_{gt})$ contains all the target molecules and template as nodes and the observed ground-truth reactions as edges. The complete bipartite graph $G_{full} = (M, T, E_{full})$ can be obtained by connecting all template $T$ nodes with the molecules node $M$. However, $G_{full}$ contains a mixture of feasible, infeasible, and invalid reactions, which should be further denoised. Naturally, our problem is transformed into obtaining the best-enhanced graph $G_{enh} = (M, T, E_{enh})$ such that $E_{gt} \subset E_{enh} \subset E_{full}$. 
 
 \textbf{Stage A}: In the first stage, we use domain knowledge and a rule-based approach to filter out edges $E_{fail}$ that generate invalid reactions from target-template subgraph mismatch or syntactically illegal structures. We obtain a potentially enhanced graph $G_{enh}^{\prime} = (M,T,E_{enh}^{\prime})$, which contains all observed ground-truth reactions and unobserved potential reactions. However, $G_{enh}^{\prime}$ still contains edges that might produce chemically infeasible reactions, creating a trade-off between template diversity recall and observed selection bias precision. 
  Naturally, this is the point where we can resort to EBM to evaluate the deviation from the observed ground-truth reaction to the unobserved enhanced reactions to obtain the best-enhanced graph. However, directly using the full set of the unobserved reaction~($>2$ million) as the surrogate for $\mathbf{x^{\prime}}$ in Eq.~\ref{eq:gradient} is still computationally infeasible. To make the training realizable and reliable, we designed a tractable subgraph-aware EBM loss to realize the training. 

 \textbf{Stage B:} We design a subgraph-aware sampling method to select the most informative subsets to build EBM loss. Specifically, for each ground-truth reaction $e_{mt} \in E_{gt}$, we adopt a k-hop reaction subgraph extraction algorithm to acquire a subgraph $G_{sub}^{m,t} \in G_{enh}^{\prime}$. This algorithm aims to extract subgraphs that contain sufficient neighborhood information in both the molecule and template dimensions to approximate the divergence of the unobserved potential reactions from the ground truth reactions. The complete algorithm is listed as Alg.~\ref{alg:khop} in the Appendix. 

 Within a selected subgraph, we can derive our EBM training objective. We omit the superscript $^{m,t}$ in the following notations for simplicity. Each subgraph $G_{sub}$ can be further divided into two counterparts: $G_{sub}^{+}$ with edges $E_{sub}^{+} = E_{sub} \cap E_{gt}$ and $G_{sub}^{-}$ with edges $E_{sub}^{-} = E_{sub} \cap E_{enh}$. We define the tractable subgraph EBM loss in Eq.~\ref{eq:loss} to push the energy score lower for the positive edges $E_{sub}^{+}$ and higher for the negative edges $E_{sub}^{-}$. Specifically, for $e_{mt}^{+} \in E_{sub}^{+}$, we have:
 
\begin{small}
\begin{equation*}
    L(\theta) = -\frac{1}{|E_{sub}^{+}|}\log\left(\frac{\exp(-F_{\theta}(e_m^+, e_t^+)/\tau)}{\sum_{e^{-}_{mt} \in E_{sub}^{-}}\exp(-F_{\theta}(e_m^-, e_t^-)/\tau)}\right)
\label{eq:loss}
\end{equation*}
\end{small}
\par 
Intuitively, each extracted subgraph $G_{sub}$ contains a set of similar reactions that reflect a selection bias towards certain types of retrosynthesis strategies. Therefore, we apply $\frac{1}{|E_{sub}^{+}|}$ as an importance weighting coefficient to alleviate the selection bias that exists in the original ground-truth distribution~\cite{importance1}. 

\textbf{Stage C:} In the denoising stage, we similarly extract the k-hop reaction subgraph for each ground-truth reaction in $E_{gt}$ and select the top-n reactions $E_{enh}^{n} \subset E_{sub}^{-}$ with the highest energy scores to form the enhanced set $E_{enh} = E_{gt} \bigcup E_{enh}^{n}$. Eventually, we can obtain the final enhanced graph $E_{enh}$ used for pre-training the downstream baseline models. The complete architecture details of the EBM and the ablation study on the different settings of $n$ are elaborated in the Appendix.

\subsection{Performance Analysis}

As shown in Table.~\ref{res:label}, there is a significant performance improvement for conceptually-enhanced models over the baselines in both ID and OOD set for retro-templates and ID set for minimal-templates, which proves that concept enhancement is effective towards countering label shifts. Nevertheless, this approach has minimal effect on the minimal-template OOD set, as the algorithm can only use retro-templates for enhancement. Among the five baselines, we discover MT demonstrates the greatest improvements, which mainly due to its ``template assembly'' capability with the enhanced dataset to further derive novel implicit retro-strategies. Specifically, we discover that MT is capable of ``assembling templates'' from the training set to generate new templates. As shown in Fig.\ref{fig:mt-generalize} in the Appendix, an unseen minimal-template in the test set can be inherently invented by combining components of different minimal-templates in the training set. Therefore, we claim that MT can potentially learn to ``invent'' unseen minimal-templates from the training data in such a manner.

\section{Conclusion}

In this study, we examined the distributional shifts in retrosynthesis prediction and proposed two model-agnostic approaches, invariant learning, and concept enhancement, to address these shifts. Furthermore, we gained insights into the impact of covariate shift and label shift on multiple baseline performances through empirical analysis and evaluation of various baseline models. We hope this work provides a foundation for understanding and addressing the distributional shifts in retrosynthesis prediction for future research. Due to most of the reaction databases are not publicly available, our benchmark dataset is focused on the open-source UPSTO dataset. Future works can extend the coverage of the reactions in the benchmark dataset by exploiting a larger private or licensed dataset to obtain a more comprehensive outcome. 

\section{Acknowledgement}

This work is sponsored by the Starry Night Science Fund at Shanghai Institute for Advanced Study~(Zhejiang University) and Shanghai AI Laboratory.

\bibliography{reference}

\appendix

\section{Algorithms}
\label{appendix:alg}

\begin{savenotes}
\begin{algorithm}[H]
   \caption{\textsc{K-hop Reaction-level Subgraph Extraction}}
   \label{alg:khop}
\begin{algorithmic}
   \STATE {\bfseries Input:} The adjacency matrix $A$ of bipartite graph $G_{enh}^{\prime}(M,T,E_{enh}^{\prime})$, a molecule-template edge $e_{mt}$ , $K \in \mathbb{Z}^{+}$
   \STATE {\bfseries Output:} the K-hop reaction-level subgraph $G_{sub}^{m,t}$ for $e_{mt}$
   \STATE Initialize $M_{rim}=\{m\}, T_{rim}=\{t\}$
   \FOR{$i=1$ {\bfseries to} $K$}
   \STATE $M_{rim}^{\prime} = \{m:m \in N_{1}(T_{rim})\ )\}$ \footnote{$N_{1}(\cdot)$ indicates the set of all first-order neighbors of the nodes in the input set.}
   \STATE $T_{rim}^{\prime} = \{t:t \in N_{1}(M_{rim})\}$
   \STATE $M_{rim} = M_{rim}^{\prime}, T_{rim} = T_{rim}^{\prime}$
   \ENDFOR
   \STATE Generate the vertex-induced subgraph $G_{sub}^{m,t}$ with vertices set $M_{rim}, T_{rim}$
\end{algorithmic}
\end{algorithm}

\end{savenotes}

\section{Additional Experiment Details}

\subsection{Granularity of Labels}
\label{appendix:granularity}

\begin{figure}[ht]
\begin{center}
\centerline{\includegraphics[width=\columnwidth]{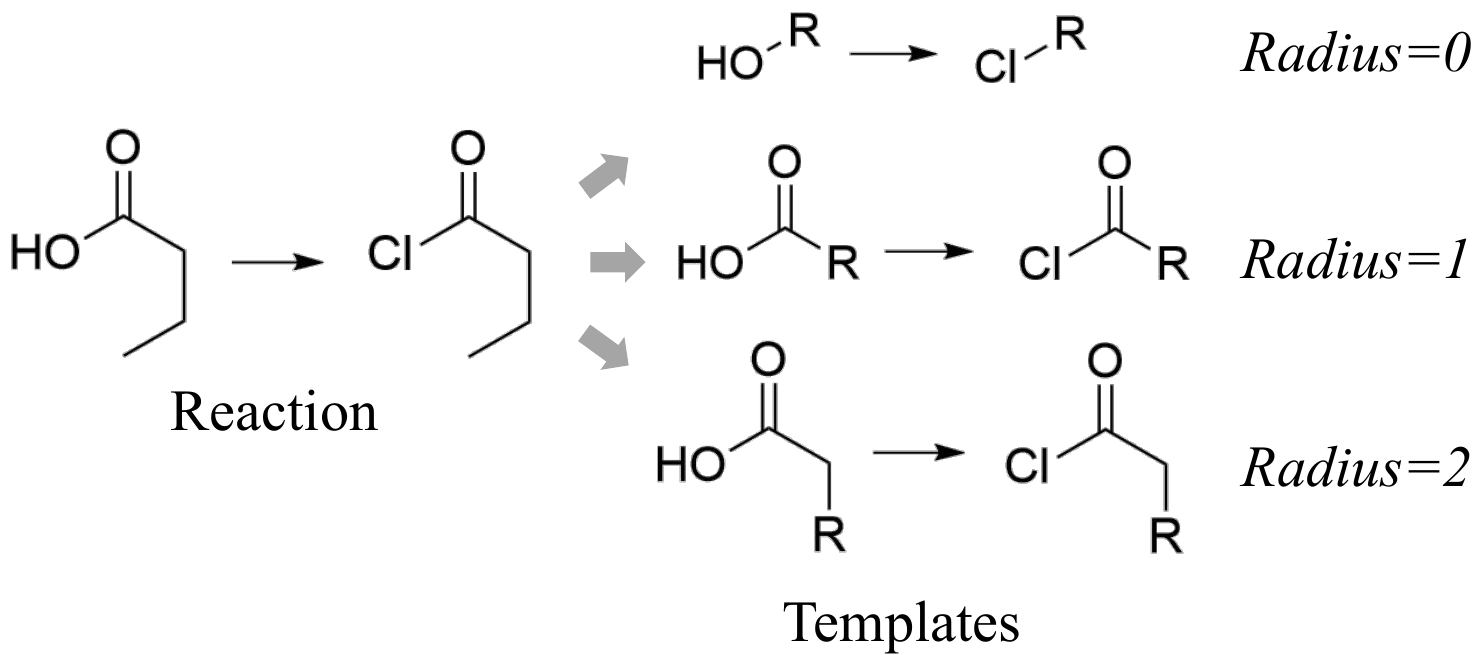}}
\caption{0, 1, and 2 steps of extrapolation from the reaction center.}
\label{fig:tpl_generation}
\end{center}
\end{figure}

 We use the reaction templates from template-based models to represent the retro-strategies for visualization. As demonstrated in Fig.~\ref{fig:tpl_generation}, we can extract reaction templates from a single-step reaction with different granularity settings, which is determined by the radius from the reaction center to the neighborhood atoms/functional groups. It is worth noting that each increment in the radius may include certain bonds or substructures (e.g., C=O) denoted chemically meaningful by experts. Often, A template class with a smaller radius can generally categorize more reaction samples. Still, it is also more prone to label shift as there is less substructure overlapping and information sharing between different classes. In contrast, a template class with a larger radius covers a smaller proportion of reaction samples but significantly increases the inter-label generalization ability. 

\begin{figure*}[htb]
\begin{center}
\centerline{\includegraphics[width=0.85\linewidth]{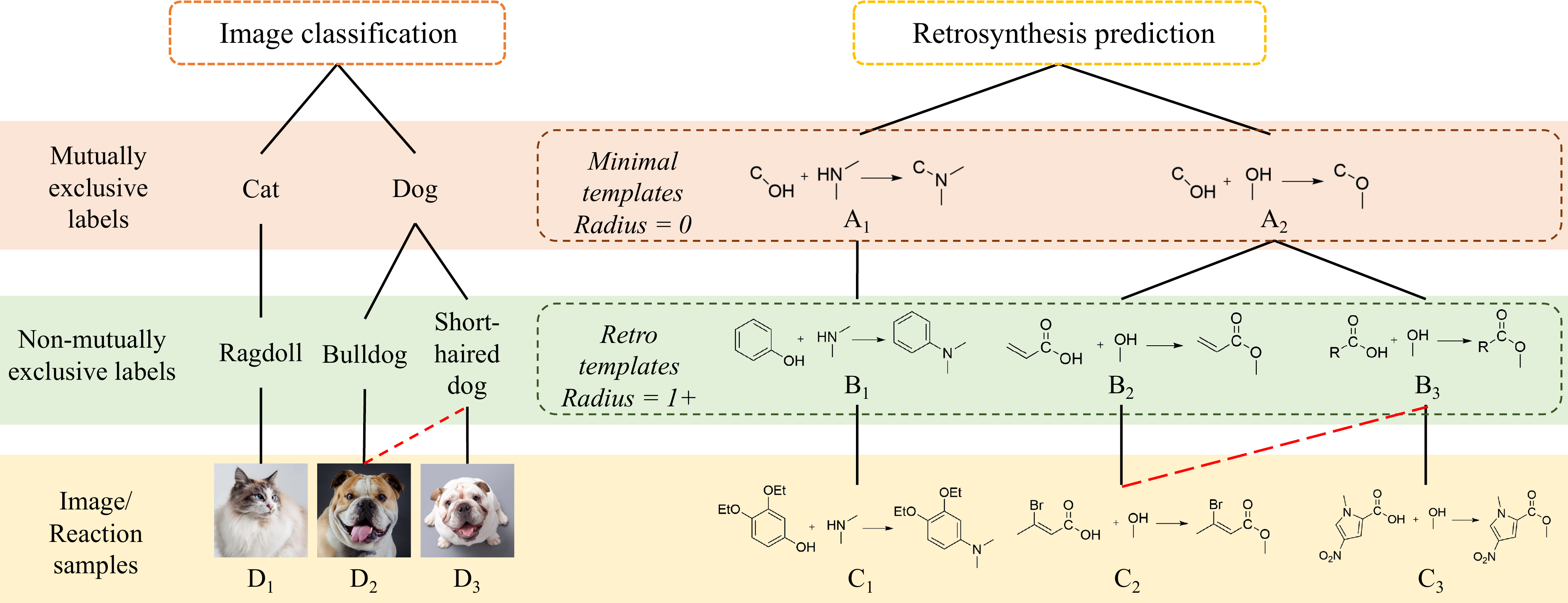}}
\vspace{-0.2cm}
\caption{minimal-templates and retro-templates. Left: In the image classification task, cats and dogs are typically regarded as mutually exclusive, while ragdolls, bulldogs, and short-haired dogs are not. Though image $D_2$ is only annotated as a bulldog, it has the potential label of a short-haired dog. Right: $A_1$ and $A_2$ are minimal-templates and $B_1$, $B_2$ and $B_3$ are retro-templates. minimal-templates and retro-templates are extracted from three reactions($C_1$, $C_2$, and $C_3$) with radius=0 and radius=1, respectively. Similar to the image $D_2$, both labels $B_2$ and $B_3$ are viable options for generating the correct reaction $C_2$, but only one template $B_2$ is exposed as the positive label in the training dataset due to non-mutual exclusivity.}
\label{fig:tpl_illustration_appx}
\end{center}
\vspace{-0.6cm}
\end{figure*}

 To better understand the retro-strategy granularity, we use an analogy between labels in image classification and template-based retrosynthesis as an intuitive interpretation. Fig.\ref{fig:tpl_illustration_appx} draws an analogy between image and reaction template labels. Generally speaking, cats and dogs are regarded as mutually exclusive labels, as no image could be labeled as cats and dogs simultaneously by definition. On the contrary, bulldogs and short-haired dogs are non-mutually exclusive due to images like $D_{2}$. In the retrosynthesis task, reaction templates could be designed to be mutually exclusive according to their radius. No reaction can simultaneously belong to multiple minimal-templates~(radius=0), but reactions, e.g., $C_{2}$, can be generated from more than one retro-templates~(radius=1), e.g., $B_2$ and $B_3$. In conclusion, a reaction can be categorized under multiple retro-template labels but only under one minimal-template label. 

\subsection{Benchmark Datasets}
\label{appendix:benchmark}

  \textbf{Label shift benchmark dataset }The OOD subset was randomly selected from the tail part of the original long-tail distribution in \texttt{USPTO-50k}. Meanwhile, it is claimed by some studies~\cite{temp1, lin2020auto} that template-free models are capable of generalizing to novel retro-strategies rather than simply recalling previously learned ones from a fixed training set, which is an advantage over template-based models~\cite{mt}. Therefore, we also deliberately ensured that the template set does \textit{not} intersect between ID and OOD subsets to investigate the ability of different models to generalize to novel retro-strategies. We begin by extracting all retro-templates and minimal-templates for each reaction in the USPTO50k dataset. Next, we randomly sample a template from the respective template set and select all reactions categorized under the sampled template to be included in our OOD subset. This process continues until the OOD test set comprises approximately 10\%~(\~5,000) of the full dataset. Subsequently, the remaining reactions are considered the ID set and randomly divided into ID train, validation, and test sets. The final ratio of the template classes between ID and OOD subsets is approximately 3:20 for minimal-templates and 9:5 for retro-templates, respectively, but can vary with different random seeds within a narrow range. Both ID subsets were divided into train/val/test sets as regular random splits, with sample size distribution for train/val/test/OOD\_test at roughly 7:1:1:1. 

  \textbf{Covariate shift benchmark dataset }To disentangle the effect of label shift and create a pure covariate shift dataset, we must ensure that at least one sample is included in the OOD subset for each template. Thus, we discard rare minimal-templates with reaction samples $<10$ ~(4472 reactions discarded). Eventually, \texttt{USPTO-50k\_M} contains $10\%$ of the samples with the most complex target molecules as the OOD subset. The rest of the samples in the ID subset are randomly split to obtain sample size distribution for train/val/test/OOD\_test at roughly 7:1:1:1.

\subsection{Baseline Methods}
\label{appendix:baselines}
 
\begin{itemize}
    \item \textit{Template-based}: Both GLN and MHN models retrosynthesis prediction as classification tasks using template-based. GLN utilizes a \textit{graph-based} representation to encode molecules and templates and applies a hierarchical joint distribution to model $P(\mathcal{T}|\mathcal{M})$. MHN utilizes a \textit{fingerprint-based} modern Hopfield network~\cite{hopfield} to associate relevant templates to product molecules. Both methods use a template retrieval-based approach that allows superior generalization across templates compared to a categorical label space such as NeuralSym~\cite{neuralsym}.
    \item \textit{Template-free}: MT models retrosynthesis as a generative neural machine translation~(NMT) problem, which uses the Transformer architecture to output reactant given a product with \textit{string-based} representation~(SMILES). The retro-strategy is implicitly learned as the mapping from the product tokens to reactant tokens. Specifically, we use the Augmented MT~\cite{augmentmt} for better performance. 
    \item \textit{Semi-template-based}: Both GraphRetro and RetroComposer exploit the \textit{graph-based} representation for molecules and approach the problem in a two-stage semi-template manner. GraphRetro contains a two-stage classification sub-task, where the first subtask is graph-edit prediction, and the second subtask is leaving-group identification. Similarly, RetroComposer divides the prediction process into template composer and reactant scoring models.
\end{itemize}

\subsection{IRM implementations}
\label{appendix:irm}

Although it is straightforward to apply the IRM regularization directly on the five baseline models, we observed that the results were not satisfactory with a naive application as shown in Tab.~\ref{res:covariate}. Specifically, we discovered that learning substructure invariance for retrosynthesis is limited to learning the invariant relationship between the substructures of the product $(\mathcal{M}_{inv})$ and the disconnection site $(\mathcal{T}_1)$, rather than other factors like leaving group selection $(\mathcal{T}_2)$. For approaches such as GLN, GraphRetro, and RetroComposer that use multi-stage losses, applying the IRM regularization to each loss term is too strong and causes an over-constrained optimization problem. In the case of MT, the problem worsens since the regularizer is applied to NLL loss for each output token, and it is more challenging to isolate the center identification stage. MT also suffers from the gap between the teacher-forcing training paradigm and the auto-regressive generation during inference~\cite{mtgap}, leading to an increased bias towards tokens that are spuriously correlated with certain retro-strategies. Thus, we design unique applications of IRM for each baseline, and the main idea is to disentangle the loss components and apply the IRM regularization to disconnection site prediction loss only. 

For the specific IRM settings, we adopt the tractable IRMv1 regularizer\cite{irmv1}:
\begin{small}
\begin{equation*}
    \min _{\Phi} \sum_{e \in \mathcal{E}_{\mathrm{tr}}}\left[\underbrace{R^e(w \cdot \Phi)}_{\text{ERM loss}}+\lambda \cdot\underbrace{\left\|\nabla_{w \mid w=\mathbf{1}} R^e(w \cdot \Phi)\right\|_2^2}_{\text{IRM penalty}}\right]
\end{equation*}
\end{small}
\par 
where $\mathcal{E}_{\mathrm{tr}}$ is the environments in the training set, $\lambda$ is the hyperparameter balancing between the predictive power over training tasks~(ERM loss) and the squared gradient norm~(IRM penalty). Importantly, the penalty term captures how much the invariant representation $\Phi$ can be improved by adjusting its scale locally. Intuitively, if the penalty term is zero across different training domains, then the "classifier" $w$ is simultaneously optimal for all of them, resulting in $w \cdot \Phi$ being an invariant predictor among the different training domains. Following our covariate data split methodology, we performed sample binning on the ID set based on the ascending size and scaffold of the target molecule to generate domain labels for IRM regularization application.

 From initial observation, we discovered that learning substructure invariance for retrosynthesis is limited to learning the invariant relationship between the substructures of the product $(\mathcal{M}_{inv})$ and the disconnection site $(\mathcal{T}_1)$, rather than other factors like leaving group selection $(\mathcal{T}_2)$. Therefore, we adjust the realization of IRM regularization in each model respectively:
\begin{itemize}
 \item In the case of MT, which utilizes a sequence-to-sequence token-wise NLL loss, different aggregation methods were attempted to convert token-wise regularization into sentence-wise regularization, which includes \textbf{simple sum}~(Eq.\ref{eq1}), \textbf{max pooling}~(Eq.\ref{eq2}), and \textbf{center-token masking}~(Eq.\ref{eq3}). Specifically:
 
    \begin{small}
    \begin{align}
        \mathcal{L}_{erm}^{i} &= \log p(t_{i}|\phi(t_{1},\dots,t_{i-1}, M)) \nonumber \\
        \mathcal{J}_{irm}^{i} &= ||\nabla_{w=1}w \cdot \mathcal{L}_{erm}^{i}||^2 \nonumber \\
        \mathcal{L}_{irm}^{sum} &=  \sum_{i=1}^{n}(\mathcal{L}_{erm} + \alpha \mathcal{J}_{irm}^{i}) \label{eq1} \\ 
        \mathcal{L}_{irm}^{max} &=  \sum_{i=1}^{n}\mathcal{L}_{erm} + \alpha \max_{i}\mathcal{J}_{irm}^{i} \label{eq2}\\ 
        \mathcal{L}_{irm}^{mask} &=  \sum_{i=1}^{n}\mathcal{L}_{erm} +  \alpha \sum_{i \in \mathcal{C}}\mathcal{J}_{irm}^{i} \label{eq3} 
    \end{align}
    \end{small}
    \par
    where $\phi(\cdot)$ is the transformer model before the token logits, $\alpha$ is the penalty weight,  and $\mathcal{C}$ comprises the reactant token indexes correlated with the disconnection site.
    \item GLN's loss function is composed of three distinct Negative Log-Likelihood~(NLL) losses for \textbf{reaction center prediction}, \textbf{template matching prediction}, and \textbf{reaction likelihood prediction}. Similarly, both GraphRetro and RetroComposer's loss function has two components for each stage (1) reaction center identification and (2) reactants structure completion. Therefore, we can directly apply the IRM regularizer to the specific loss for identifying the disconnection site. For MHN, however, the model calculates the NLL of template-product relevance directly from the template and product fingerprints, making it impossible to disentangle the center identification loss without re-designing the model architecture. 
\end{itemize}

\subsection{Visualization}
\label{appendix:visualize}


\begin{figure*}[ht]
\begin{center}
\centerline{\includegraphics[width=1.0\linewidth]{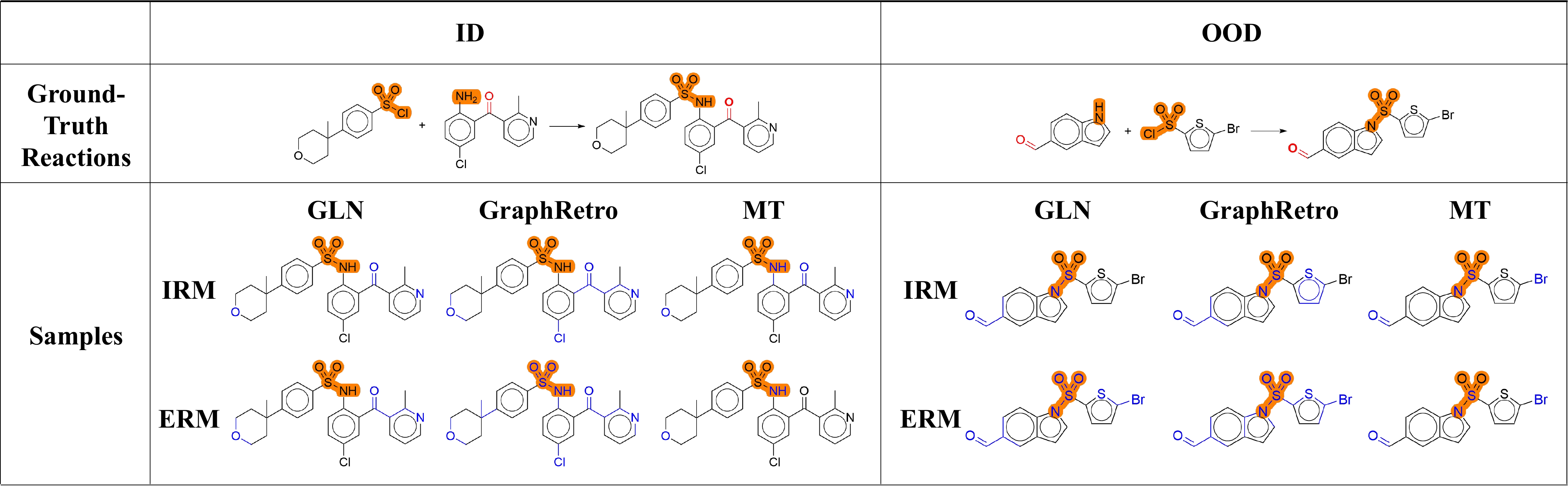}}
\caption{The visualization map in two reactions selected from the ID and OOD test set in $\mathtt{USPTO50K\_M}$, respectively. The ground-truth disconnection site is marked in orange, and the relevant substructures are annotated in red by chemists. On the heat map of the molecules, the salient substructures are highlighted in blue under different settings.}
\label{fig:visualization_analysis}
\end{center}
\vskip -0.1in
\end{figure*}

To comprehensively analyze how IRM reshapes the retrosynthesis models in capturing the substructure invariance in reaction center identification, we apply different axiomatic attribution methods to acquire an explainable interpretation. Specifically, we use the atom-aggregated attention weights visualization for MT and the gradient activation map~\cite{explaingraph} for another two graph-based models. 

Fig.~\ref{fig:visualization_analysis} illustrates the visualization and the attribution annotated by chemists. Two GT reactions shown are condensation reactions resulting in the synthesis of sulfonamide. The substructures highlighted in red by the chemists are highly related to selecting the orange disconnection site, the sulfonamide functional group~(N-S(=O)(=O)), which reflects the selectivity rules in the laboratory experiments. In general, we found that using IRM regularization results in a reduction of spurious correlation on variant substructures $\mathcal{M}_{var}$ and an increased convergence towards the annotated invariant substructures $\mathcal{M}_{inv}$. Additionally, models trained under ERM tend to assign a more evenly distributed weight to different substructures, but a majority of these additional atoms are not considered as an invariant feature for predicting the specific reaction center, as seen in the visual analysis in Fig.\ref{fig:visualization_analysis}.

\begin{figure}[htb]
\vspace{-0.1cm}
\begin{center}
\centerline{\includegraphics[width=\linewidth]{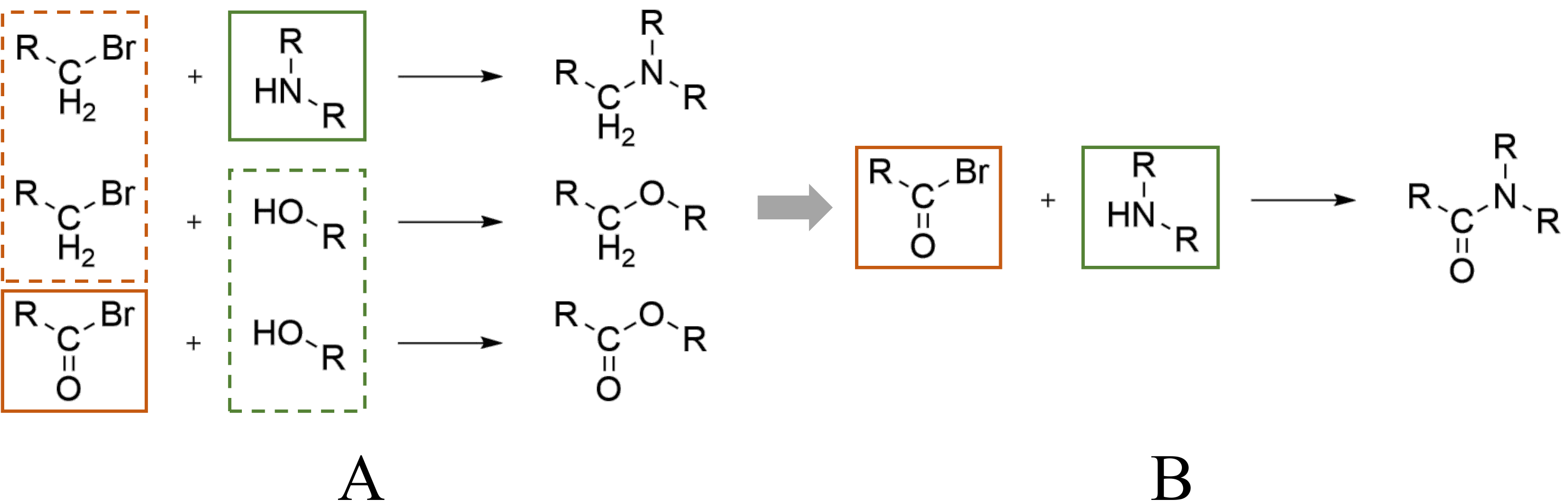}}
\caption{A demonstration of MT's ``template assembly'' capability from three minimal-templates(A) in the training set to create one unseen minimal-template(B) in the test set.}
\label{fig:mt-generalize}
\end{center}
\vspace{-0.6cm}
\end{figure}

\begin{figure*}[ht]
\begin{center}
\centerline{\includegraphics[width=\linewidth]{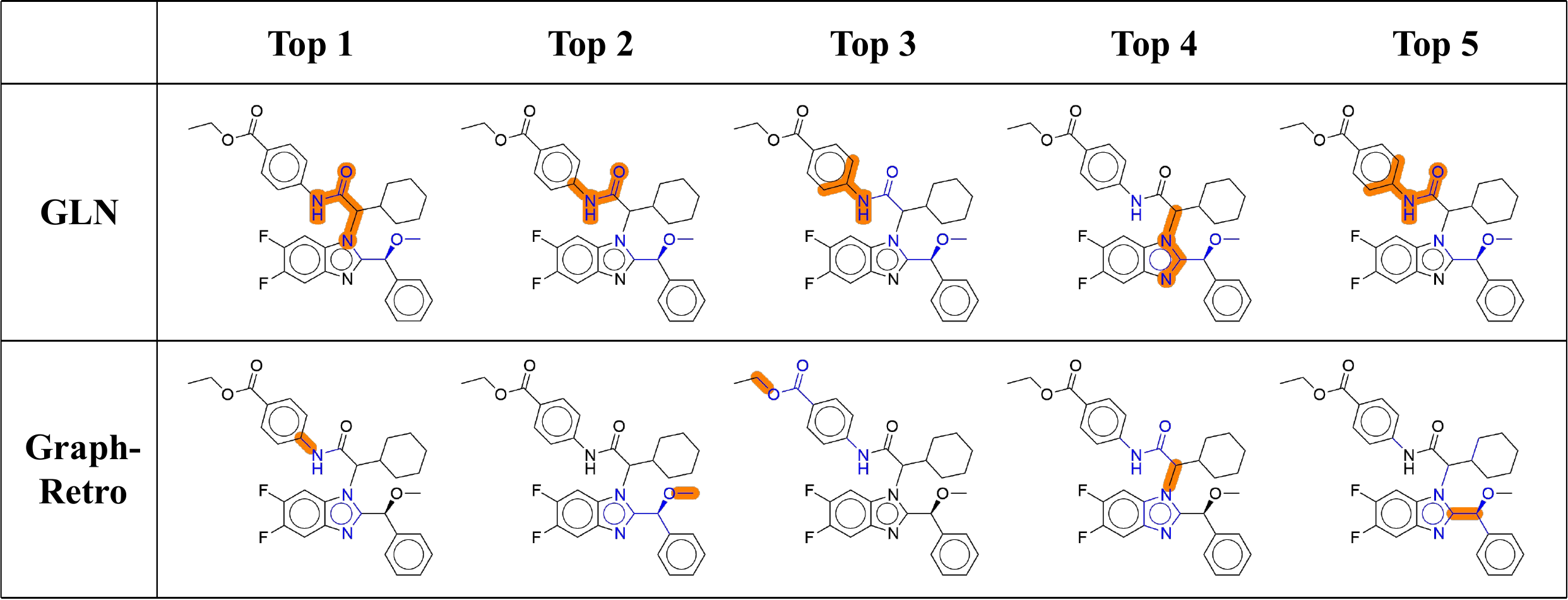}}
\caption{Top-5 predictions visualization for GLN and GraphRetro. The orange color highlights the predicted reaction center regions(GLN) or atom/bond-edit location(GraphRetro), and the blue color indicates the attributions from chemists annotations. }
\label{fig:topk}
\end{center}
\vskip -0.1in
\end{figure*}

We also provided additional visualization to analyze the difference in the graph representation between GLN and GraphRetro. As shown in Fig.\ref{fig:topk}, the saliency map of GLN's top-5 predictions is largely consistent, whereas GraphRetro's top-5 predictions differ significantly. We conclude that this phenomenon results from different architectures used for predicting the disconnection site. GLN outputs a GNN readout graph embedding and combines it with different centers in templates to output logits, while GraphRetro uses the individual edge/atom-wise embedding after the graph convolution and directly predicts if an edit can be made at each possible location.

\begin{figure}[htb]
\begin{center}
\centerline{\includegraphics[width=\linewidth]{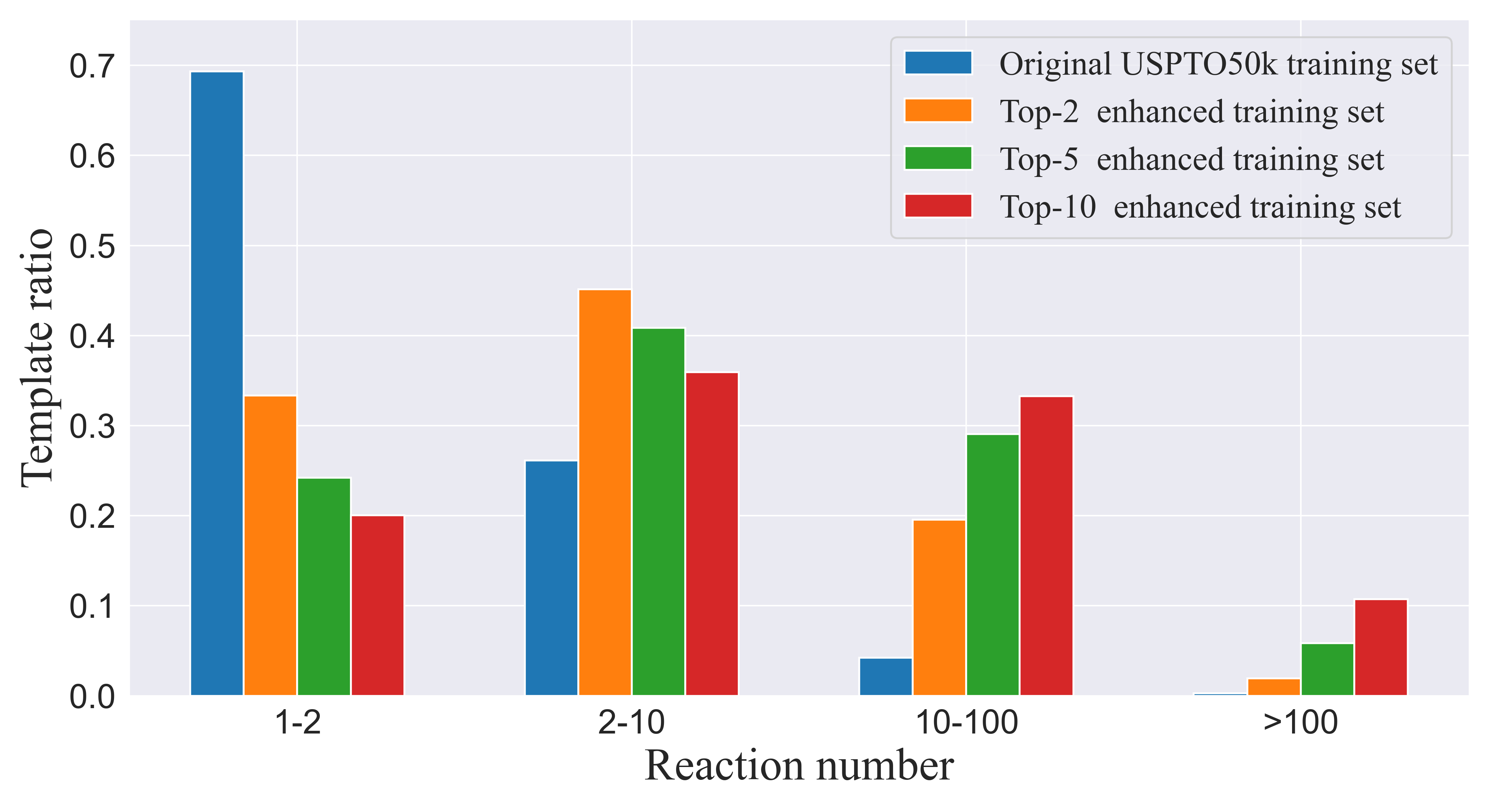}}
\caption{The template distribution with respect to the number of samples.}
\label{fig:tpl-distributions}
\end{center}
\vskip -0.1in
\end{figure}

\begin{table}[htb]
\begin{center}
\begin{small}
\begin{sc}
\resizebox{\columnwidth}{!}{
\begin{tabular}{l|ccccccr}
\toprule
mol-size  & GLN\_tp & GLN\_all  & MT\_max & MT\_sum & G\_Retro\_all & R\_composer\_all \\
\midrule
ID Top-1   & 53.5\% & 53.7\% & 50.9\% & 51.9\% & 53.7\% & 53.9\% \\
ID Top-3   & 68.2\% & 69.2\%  & 74.5\% & 76.1\% & 70.2\% & 78.8\% \\
ID Top-5  & 76.1\% & 76.7\% & 79.7\% & 81.3\% & 73.7\% & 85.3\%  \\
ID Top-10  & 84.1\% & 85.3\% & 82.6\% & 85.2\% & 76.9\% & 89.6\%\\
\midrule
OOD Top-1   & 36.9\% & 38.0\% & 29.7\%  & 30.0\% & 37.7\%  & 40.1\%\\
OOD Top-3   & 49.9\% & 50.9\% & 46.5\% & 47.5\% & 56.9\% & 66.4\%\\
OOD Top-5  &  58.5\% & 59.2\% & 53.9\% & 54.6\%  & 64.8\%  & 74.3\%\\
OOD Top-10  & 70.1\% & 71.1\% & 57.7\% & 60.4\%  & 71.1\%  & 82.8\%\\
\bottomrule
\toprule

mol-scaffold  & GLN\_tp & GLN\_all  & MT\_max & MT\_sum & G\_Retro\_all & R\_composer\_all \\
\midrule
\midrule
ID Top-1   & 54.2\% &  55.2\% & 50.1\% & 51.2\% & 55.4\%  & 50.4\%\\
ID Top-3   & 69.2\% & 69.9\% &  74.2\% & 75.6\% &69.5\%  & 77.1\% \\
ID Top-5  & 76.4\% & 77.2\% & 79.3\% & 81.6\%  &73.2\%  & 83.6\%\\
ID Top-10  & 84.7\% & 85.1\% & 82.3\% & 83.7\%  &76.2\%  & 88.8\%\\
\midrule
OOD Top-1   & 37.9\% & 39.0\% & 37.4\% & 38.6\% &39.2\%  & 39.9\%\\
OOD Top-3   & 52.3\% & 52.9\% & 57.9\% & 59.0\% &57.6\%  & 64.6\%\\
OOD Top-5  &  60.8\% & 61.2\% & 64.3\% & 65.6\% &64.9\%  & 74.5\%\\
OOD Top-10  & 72.1\% & 72.4\% & 69.5\% & 70.2\% & 71.5\%  & 82.2\%\\
\bottomrule
\end{tabular}
}
\end{sc}
\end{small}
\end{center}
\vskip -0.1in
\caption{Ablation study of using IRM regularizer for different loss. GLN: cp = reaction center prediction, tp = template matching prediction, rc = reaction likelihood prediction, all = applying to all three losses. MT: max = token with maximum loss, sum = sum all tokens. G\_Retro: all = graph edit + synthon completion. R\_composer: all = template composer + reactant scoring}
\vskip -0.1in
\label{res:cov-ablation}
\end{table}

\subsection{Details of Concept Enhancement}
\label{appendix:EBM}

\subsection{Concept Enhancement Model Architecture}

The input feature for atoms and bonds in the graph is listed. All raw features are one-hot encoded depending on feature size.

\begin{table}[htb]
\begin{center}
\begin{small}
\resizebox{\columnwidth}{!}{
\begin{tabular}{|c|c|c|}
\hline
\text { Node Feature } & \text { Description } & \text { Feature Size } \\
\hline \text { Atom category } & \text { Category of atom (e.g. C, N, O) } & \text{65} \\
\text { Atom degree } & \text { Number of bonds the atom has } & \text{10} \\
\text { Formal charge } & \text { Formal charge assigned to atom } & \text{5} \\
\text { Valency } & \text { Explicit valency of the atom } & \text{7} \\
\text { Hydrogen counts} & \text { Number of bonded Hydrogen atoms } & \text{5} \\
\text { Aromaticity } & \text { Whether atom is part of a aromatic ring } & \text{1} \\
\text { Hybridization } & \text { Hybridization of the atom } & \text{5} \\
\hline

\text { Edge Feature } & \text { Description } & \text { Feature Size } \\
\hline
\text { Bond type } & \text { Single, double, triple, aromatic } & \text{4} \\
\text { Bond conjugation } & \text { Whether bond is conjugated } & \text{1} \\
\text { Bond in ring } & \text { Whether bond is part of an aromatic ring } & \text{1} \\
\hline
\end{tabular}
}
\end{small}
\end{center}
\end{table}

The EBM architecture design uses two separate graph encoders to encode the molecular and subgraph structures in templates, respectively. We use \textsc{average} pooling to aggregate into a single embedding for multiple structures in the reactant side of the template.  Specifically, we use similar Message Passing Neural Network~(MPNN) structures for GNN implementation as to works~\cite{graphretro, gln} with depth=10 with hidden layer dimension=256. Dropout with rate=0.1 is applied to the embedding layers and MLP layers. 

We use the training set of the retro-templates split to generate the bipartite graph. The entire enhanced bipartite graph after domain knowledge filtering contains 34750 molecule nodes, 6788 template nodes, and 2018153 edges.

\subsection{Training and Optimization}

We use $k=1$ as the k-hop reaction-level subgraph extraction for creating the training loss. For extremely large subgraphs, the maximum number of negative samples $m$ is limited to 100 or 200 as the cut-off threshold. We prioritize the negative samples with the same target molecules as the positive samples during cut-off.

We use \textsc{PyTorch} for implementation, \textsc{RDKit}~\cite{rdkit} for feature extraction of chemical structures, and \textsc{RdChiral}~\cite{rdchiral} for templates-related operations. The Adam optimizer with an initial learning of 0.001 was used to optimize the model. Training is done on two RTX3090 GPUs with batch size 16 for $m=100$ and batch size 8 for $m=200$. 

\begin{figure*}[htb]
\begin{center}
\centerline{\includegraphics[width=\linewidth]{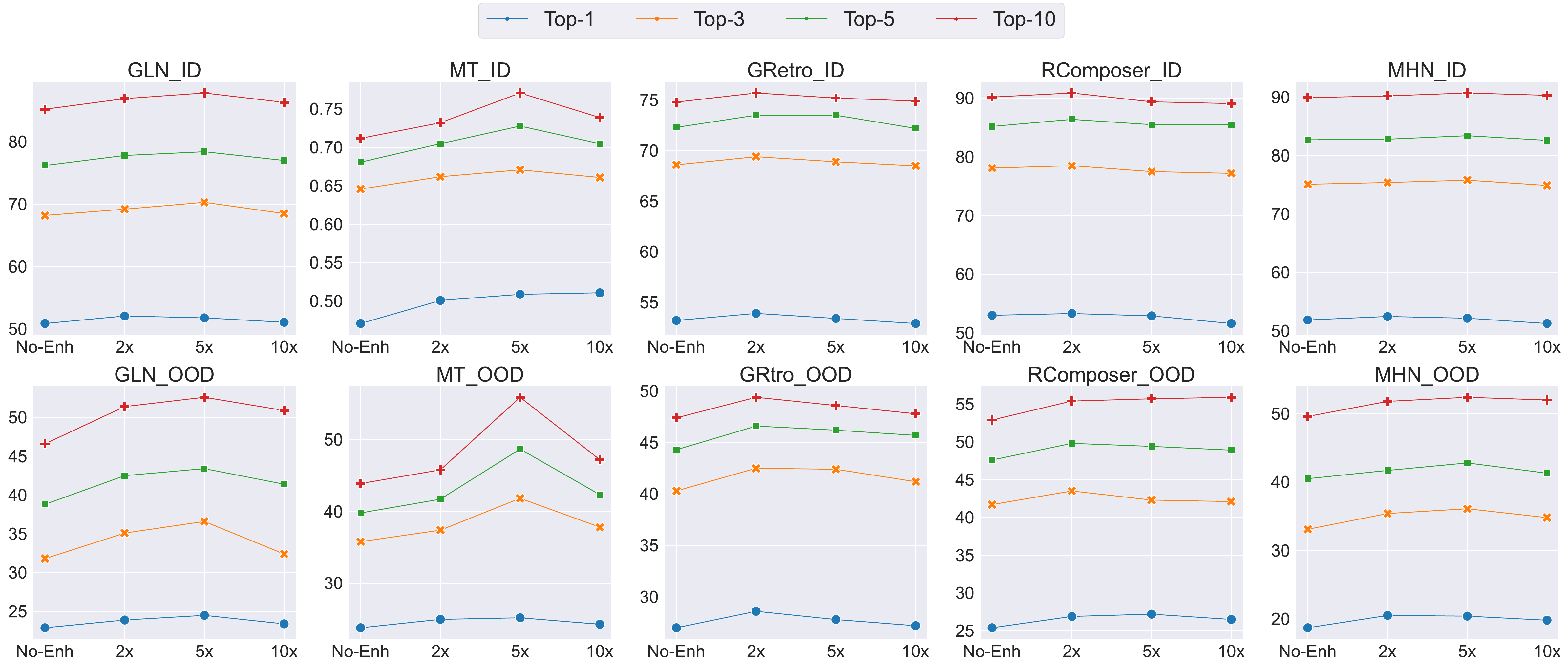}}
\caption{Top-k accuracy for baselines under the original training set and $n=2$, $n=5$, and $n=10$ times concept-enhancement.}
\label{fig:ablation_n}
\end{center}
\vskip -0.1in
\end{figure*}

\end{document}